\documentclass[11pt]{article}
\usepackage{coling2020}
\usepackage{times}
\usepackage{amssymb}
\usepackage{url}
\usepackage{enumitem}
\usepackage{hyperref}
\usepackage{latexsym}
\usepackage{adjustbox}
\usepackage{booktabs}
\usepackage[title]{appendix}

\usepackage{todonotes}

\usepackage{microtype}

\colingfinalcopy 

\title{Tackling Online Abuse: A Survey of Automated Abuse Detection Methods}

\author{Pushkar Mishra$^\bigstar$,\,\, Helen Yannakoudakis$^\spadesuit$,\,\, Ekaterina Shutova$^\clubsuit$\\
  $^\bigstar$ Facebook AI, London, United Kingdom\\
  $^\spadesuit$ Department of Informatics, King's College London, United Kingdom\\
  $^\clubsuit$ Institute for Logic, Language and Computation, University of Amsterdam, The Netherlands\\
  \normalsize{{\tt pushkarmishra@fb.com, helen.yannakoudakis@kcl.ac.uk, e.shutova@uva.nl}}
}

\date{}

\begin{document}
\maketitle
\begin{abstract}
Abuse on the Internet represents an important societal problem of our time. Millions of Internet users face harassment, racism, personal attacks, and other types of abuse on online platforms. The psychological effects of such abuse on individuals can be profound and lasting. Consequently, over the past few years, there has been a substantial research effort towards automated abuse detection in the field of natural language processing (NLP). In this paper, we aim to present a comprehensive view of the field, hence providing a platform for further development of this area. We first describe the existing datasets and review the computational approaches to abuse detection, analyzing their strengths and weaknesses. We then discuss the main trends that emerge, highlight the challenges that remain, and outline possible solutions. Finally, taking note of the developments in the field, we propose guidelines for ethics and explainability.
\end{abstract}

\section{Introduction}
\label{intro}
With the advent of social media, anti-social and abusive behavior has become a prominent occurrence online. Undesirable psychological effects of abuse on individuals make it an important societal problem of our time. Munro \shortcite{munro2011} studied the ill-effects of online abuse on children, concluding that children may develop depression, anxiety, and other mental health problems as a result of their encounters online. \textit{Pew Research Center}, in its latest report on online harassment \cite{pew}, revealed that $40\%$ of adults in the United States have experienced abusive behavior online, of which $18\%$ have faced severe forms of harassment, e.g., that of sexual nature. The report goes on to say that harassment need not be experienced first-hand to have an impact: $13\%$ of American Internet users admitted that they stopped using an online service after witnessing abusive and unruly behavior of their fellow users. These statistics stress the need for automated abuse detection and moderation systems. Therefore, in the recent years, a new research effort on abuse detection has sprung up in NLP.

That said, the notion of abuse has proven elusive and difficult to formalize. Different norms across (online) communities can affect what is considered abusive \cite{chandrasekharan2018internet}. In the context of natural language, \textit{abuse} is a term that encompasses many different types of fine-grained negative expressions. For example, Nobata et al. \shortcite{nobata} use it to collectively refer to hate speech, derogatory language and profanity, while Mishra et al. \shortcite{mishra} use it to discuss racism and sexism. The definitions for different types of abuse tend to be overlapping and ambiguous. However, regardless of the specific type, we define abuse as \textit{any expression that is meant to denigrate or offend a particular person or group}.
Taking a course-grained view, Waseem et al. \shortcite{W17-3012} classify abuse into broad categories based on \textit{explicitness} and \textit{directness}. \textit{Explicit} abuse comes in the form of expletives, derogatory words or threats, while \textit{implicit} abuse has a more subtle appearance characterized by the presence of ambiguous terms and figures of speech such as metaphor or sarcasm. \textit{Directed} abuse targets a particular individual as opposed to \textit{generalized} abuse, which is aimed at a larger group such as a particular gender or ethnicity.

This categorization exposes some of the intricacies that lie within the task of automated abuse detection. While directed and explicit abuse is relatively straightforward to detect for humans and machines alike, the same is not true for implicit or generalized abuse. This is illustrated in the works of Dadvar et al. \shortcite{davdar} and Waseem and Hovy \shortcite{waseem_hovy}: Dadvar et al. observed an inter-annotator agreement of $93\%$ on their cyber-bullying dataset. Cyber-bullying is a classic example of directed and explicit abuse since there is typically a single target who is harassed with personal attacks. On the other hand, Waseem and Hovy noted that $85\%$ of all the disagreements in annotation of their dataset occurred on the sexism class. Sexism is typically both generalized and implicit.

This paper aims to provide a comprehensive view of the field of automated abuse detection. We make various contributions that differ from \textit{traditional} surveys of the field \cite{schmidt_wiegand,fortuna_survey,salminen2018anatomy,castelle2018linguistic}. Firstly, we present a review of the commonly-used datasets. We then discuss the various methods for abuse detection that have been investigated by the NLP community, including ones based on neural networks which previous surveys have omitted. Next, we summarize the main trends that emerge, highlight the challenges that remain, and outline possible solutions. Lastly, taking note of the direction of developments, we propose guidelines for ethics and explainability that align with the aforementioned categorization of abuse per explicitness and directness.


\section{Annotated datasets}
Supervised learning approaches to abuse detection require annotated datasets for training and evaluation purposes. To date, several manually annotated datasets have been made available by researchers. Before describing the commonly-used ones,\footnote{Some datasets are directly described in the later sections where methods applied on them are discussed. Additionally, in the appendix, we also provide summaries of the datasets that are publicly available along with links to them.} we highlight the two respects in which these datasets differ:
\vspace{-2mm}

\begin{itemize}
    \item\textit{Source}: the platform from which the data samples were collected. For example, data samples can be posts from \textit{Reddit} or tweets from Twitter. Source governs many properties of the dataset such as linguistic style and structure, level of grammatical correctness, extent of (deliberate) obfuscation of words, etc. Essentially, source affects both explicitness and directness of the abusive samples in it.
    \vspace{-1.35mm}
    \item\textit{Composition}: the composition of a dataset is governed by the nature of data samples it contains. Most datasets are annotated for or compiled to cover only certain subset of types of abuse, e.g., racism and sexism, or personal attack and racism, or hate speech and profanity.
\end{itemize}

\noindent
\textbf{Early datasets.} The earliest dataset published in this field was from Spertus \shortcite{smokey}. It consisted of $1,222$ private messages written in English taken from web-masters of controversial web resources such as \textit{NewtWatch}. These messages were marked as \textit{flame} (containing insults or abuse; $7.5\%$), \textit{maybe flame} ($13\%$), or \textit{okay} ($79.5\%$). We refer to this dataset as \textsc{data-smokey}. 
Yin et al. \shortcite{Yin09detectionof} constructed three English datasets and annotated them for \textit{harassment}, which they defined as ``systematic efforts by a user to belittle the contributions of other users". The samples were taken from three social media platforms: \textit{Kongregate} ($4,802$ posts; $0.87\%$ harassment), \textit{Slashdot} ($4,303$ posts; $1.4\%$ harassment), and \textit{MySpace} ($1,946$ posts; $3.3\%$ harassment). We refer to the three datasets jointly as \textsc{data-harass}.

\noindent
\textbf{Yahoo! as source.} Many datasets have been compiled using samples taken from portals of \textit{Yahoo!}, specifically the \textit{News} and \textit{Finance} ones. Djuric et al. \shortcite{djuric} created a dataset of $951,736$ user comments in English from the \textit{Yahoo! Finance} website that were editorially labeled as \textit{hate speech} ($5.9\%$) or \textit{clean} (\textsc{data-yahoo-fin-dj}). Nobata et al. \shortcite{nobata} produced four more datasets with comments from \textit{Yahoo! News} and \textit{Yahoo! Finance}, each labeled \textit{abusive} or \textit{clean}: 1) \textsc{data-yahoo-fin-a}: $759,402$ comments, 7.0\% abusive; 2) \textsc{data-yahoo-news-a}: $1,390,774$ comments, 16.4\% abusive; 3) \textsc{data-yahoo-fin-b}: $448,436$ comments, 3.4\% abusive; and 4) \textsc{data-yahoo-news-b}: $726,073$ comments, 9.7\% abusive.

\noindent
\textbf{Twitter as source.} Several groups have investigated abusive language in Twitter. Waseem and Hovy \shortcite{waseem_hovy} 
created a corpus of $16,907$ tweets, each annotated as one of \textit{racism} ($11.7\%$), \textit{sexism}, ($20.0\%$) or \textit{neither} (\textsc{data-twitter-wh}). We note that although certain tweets in the dataset lack explicit abusive traits (e.g., \textit{@Mich\_McConnell Just ``her body'' right?}), they have nevertheless been marked as racist or sexist as the annotators took the wider discourse into account; however, such discourse information is not preserved in the dataset. Inter-annotator agreement was reported at $\kappa = 84\%$, with a further insight that $85\%$ of all the disagreements occurred on the sexism class alone.
Waseem \shortcite{waseem} later released a dataset of $6,909$ tweets annotated as \textit{racism} ($1.41\%$), \textit{sexism} ($13.08\%$), \textit{both} ($0.70\%$), or \textit{neither} (\textsc{data-twitter-w}). \textsc{data-twitter-w} and \textsc{data-twitter-wh} have $2,876$ tweets in common. 
It should, however, be noted that the inter-annotator agreement between the two datasets is low (mean pairwise $\kappa = 14\%$) \cite{waseem}. Davidson et al. \shortcite{davidson} 
created a dataset of approximately $25k$ tweets, manually annotated as one of \textit{racist} ($5\%$), \textit{offensive but not racist} ($76\%$), or \textit{clean} ($19\%$). 
We note, however, that their data sampling procedure relied on the presence of certain abusive words and, as a result, the distribution of classes does not follow a real-life distribution. 
Recently, Founta et al. \shortcite{founta} crowd-sourced a dataset (\textsc{data-twitter-f}) of $80k$ tweets, of which $59\%$ were rated \textit{normal}, $22.5\%$ \textit{spam}, $7.5\%$ \textit{hateful}, and $11\%$ \textit{abusive}. The \textit{OffensEval} 2019 shared task used a recently released dataset of $14,100$ tweets \cite{OLID}, each hierarchically labeled as: offensive ($33\%$) or not, offense is targeted ($29\%$) or not, and whether the target is an individual ($17.8\%$), a group ($8.2\%$) or otherwise ($3\%$).

\noindent
\textbf{Other English sources.} Wulczyn et al. \shortcite{wulczyn} crowd-sourced annotations for English comments from \textit{Wikipedia'}s \textit{talk pages} and released three datasets: one focusing on personal attacks ($115,864$ comments; $11.7\%$ abusive), one on aggression ($115,864$ comments), and one on toxicity ($159,686$ comments; $9.6\%$ abusive) (\textsc{data-wiki-att}, \textsc{data-wiki-agg}, and \textsc{data-wiki-tox} respectively). \textsc{data-wiki-agg} contains the exact same comments as \textsc{data-wiki-att} but annotated for aggression -- the two datasets show a high correlation in the nature of abuse (\textit{Pearson's} $r=0.972$).
Gao and Huang \shortcite{gao2017detecting} released a dataset of $1,528$ Fox News user comments (\textsc{data-fox-news}) annotated as \textit{hateful} ($28.5\%$) or \textit{non-hateful}. The dataset preserves context information for each comment, including user's screen-name, all comments in the same thread, and the news article for which the comment is written.
Salminen et al. \shortcite{salminen2018anatomy} sourced $137k$ comments from under videos posted on YouTube and Facebook (\textsc{data-ytube-fb}). They self-annotated the corpus per a fine-grained hierarchical taxonomy consisting of 13 main categories and 16 sub-categories that cover both nature of abuse (e.g., humiliation) as well as targets (e.g., religion).

\noindent
\textbf{Non-English datasets.} Some researchers investigated abuse in languages other than English. Van Hee et al. \shortcite{vanhee} gathered $85,485$ Dutch posts from \textit{ask.fm} to form a dataset on cyber-bullying (\textsc{data-bully}; $6.7\%$ cyber-bullying cases). 
Pavlopoulos et al. \shortcite{pavlopoulos-emnlp} released a dataset of ca. $1.6M$ comments in Greek provided by the news portal \textit{Gazzetta} (\textsc{data-gazzetta}). The comments were marked as \textit{accept} or \textit{reject}, and are divided into 6 splits with similar distributions (the training split is the largest one: $66\%$ accepted and $34\%$ rejected comments). 
As part of the \textit{GermEval} shared task on identification of offensive language in German tweets \cite{wiegand2018overview}, a dataset of $8,541$ tweets was released, of which $21\%$ were labeled as \textit{abuse}, $11.4\%$ as \textit{insult}, $1.39\%$ as \textit{profanity}, and $66.16\%$ as \textit{other}. 
Around the same time, $15k$ Facebook posts and comments, each in Hindi (in both Roman and Devanagari script) and English, were released (\textsc{data-facebook}) as part of the \textit{COLING 2018} shared task on aggression identification \cite{kumar2018benchmarking}. $35.3\%$ of the comments were \textit{covertly aggressive}, $22.8\%$ \textit{overtly aggressive} and $41.9\%$ \textit{non-aggressive}. We note, however, that some issues were raised by the participants regarding the quality of the annotations. The \textit{HatEval} 2019 shared task focused on detecting hate speech against immigrants and women using a dataset of $5k$ tweets in Spanish and $10k$ in English annotated hierarchically as hateful or not; and, in turn, as aggressive or not, and whether the target is an individual or a group.

\vspace{1mm}
\noindent
\textbf{Remarks.} In their study, Ross et al. \shortcite{ross} stressed the difficulty of reliably annotating abuse, which stems from multiple factors such as the lack of \textit{standard} definitions for the myriad types of abuse, differences in annotators' cultural background and experiences, and ambiguity in the annotation guidelines. That said, Waseem et al. \shortcite{W17-3012} and Nobata et al. \shortcite{nobata} observed that annotators with prior expertise provide good-quality annotations with high levels of agreement amongst themselves. That aside, most datasets contain discrete labels only; abuse detection systems trained on such datasets would be deprived of the notion of \textit{severity}, which is vital in real-world settings. In fact, all existing datasets cover only a subset of abuse types. Moreover, recent studies have also found substantial and systematic bias in existing datasets, which they primarily attributed to the sampling strategy, e.g., topic or word-based \cite{wiegand2019detection}.


\section{Feature engineering based abuse detection}
The first documented method for abuse detection was that of Spertus \shortcite{smokey} who hand-crafted rules over texts to generate feature vectors for learning. Since then, several methods have been proposed that rely on manual \textit{feature engineering}. Such feature engineering happens on two fronts, either on the text of the sample (\textit{textual}) or on the user(s) who created or interacted with the sample (\textit{social}).

\noindent
\textbf{Textual feature engineering.} This kind of feature engineering models directed and explicit traits of abuse within samples. Researchers have adopted two approaches to textual feature engineering: hand-crafted rules cum lexicon-based approach and computational approach. The former includes features extracted from text based on linguistic rules (e.g., text contains the pronoun \textit{you} followed by profanity) or some curated lexicon of abusive words and expressions (e.g., \href{https://hatebase.org}{\textit{hatebase.org}}. The latter on the other hand includes bag-of-words (\textsc{bow}) counts, \textsc{tf-idf} weighted features, features based on similarity clustering, etc. Both approaches are summarized in Table \ref{tab:text_feature_engg}. 

\begin{table}[t!]
    \small
    \centering
    \begin{tabular}{c|c}
    \toprule
        \textbf{Rules and Lexicon-based} &  \textbf{Computational}\\
        \midrule
        \multicolumn{1}{p{7.5cm}|}{
        Spertus \shortcite{smokey}:
        \vspace{-2.5mm}
        \begin{itemize}
            \item Employed 47 hand-crafted linguistic rules to extract binary feature vectors for feeding to a decision tree generator to train a classification model.
            \vspace{-2.5mm}
            \item $64\%$ recall on the \textit{flame} and $98\%$ on the \textit{non-flame} messages in the test set of the \textsc{data-smokey} dataset.
        \end{itemize}
        \vspace{-2.5mm}
        Razavi et al. \shortcite{razavi}:
        \vspace{-2.5mm}
        \begin{itemize}
            \item First to propose a lexicon-based method for abuse detection whereby they constructed an \textit{insulting and abusing language} dictionary of words and phrases, with each entry having an associated weight indicating its abusive impact; Njagi et al. \shortcite{gitari} also adopted such a lexicon-based approach later on.
            \vspace{-2.5mm}
            \item $96\%$ accuracy on a dataset of $1,525$ messages ($32\%$ \textit{flame} and the rest \textit{okay}) extracted from the \textit{Usenet} newsgroup and the employee conversation threads of the \textit{Natural Semantic Module} company.
        \end{itemize}
        \vspace{-2.5mm}
        Wiegand et al. \shortcite{wiegand}:
        \vspace{-2.5mm}
        \begin{itemize}
            \item Automated framework for creating lexicons whereby they first constructed a \textit{base lexicon} of \textit{negative polar expressions} with 33\% annotated as abusive, and then trained an SVM on multiple hand-crafted linguistic and semantic features which yielded an F$_1$ of $81.6\%$ on the base lexicon; they applied this SVM to a set of unlabeled negative polar expressions to expand the base lexicon with more abusive expressions.
            \vspace{-2.5mm}
            \item $63.8\%$ F$_1$ on \textsc{data-twitter-wh} and $78.4\%$ F$_1$ on \textsc{data-wiki-att} with an \textsc{svm} trained on hand-crafted rank features derived from the expanded lexicon.
        \vspace{-5mm}
        \end{itemize}
        }
        & 
        \multicolumn{1}{|p{7.5cm}}{
        Yin et al. \shortcite{Yin09detectionof}:
        \vspace{-2.5mm}
        \begin{itemize}
            \item Extracted local features (\textsc{tf-idf} weights of words), sentiment-based features (\textsc{tf-idf} weights of foul words and pronouns), and also contextual features (e.g., similarity of a post to its neighboring posts) which captured aspects not covered by former two.
            \vspace{-6mm}
            \item $31.3\%$ F$_1$, $29.8\%$ F$_1$ and $48.1\%$ F$_1$ respectively on the \textsc{data-harass} datasets, i.e., \textit{MySpace}, \textit{Slashdot}, and \textit{Kongregate}, with a linear \textsc{svm}.
        \end{itemize}
        \vspace{-2.5mm}
        Sood et al. \shortcite{sood2012}:
        \vspace{-2.5mm}
        \begin{itemize}
            \item Word bi-gram features combined with a word-list baseline utilizing a \textit{Levenshtein distance} based heuristic.
            \vspace{-6mm}
            \item $63\%$ F$_1$ using an \textsc{svm} on a dataset of 6,354 comments from a social news site labeled \textit{profanity} (9\%) or not.
        \end{itemize}
        \vspace{-2.5mm}
        Warner and Hirschberg \shortcite{warner}:
        \vspace{-2.5mm}
        \begin{itemize}
            \item Framed their task as word-sense disambiguation, i.e., whether a term carried an \textit{anti-semitic} sense or not, and  employed a template-based strategy alongside \textit{Brown clustering} to extract surface-level \textsc{bow} features.
            \vspace{-2.5mm}
            \item $63\%$ F$_1$ using an \textsc{svm} on a dataset of 1000 paragraphs annotated as \textit{anti-semitic} (9\%) or not.
        \end{itemize}
        \vspace{-2.5mm}
        Dinakar et al. \shortcite{dinakar2011modeling}, Burnap et al. \shortcite{burnap}, Hee et al. \shortcite{vanhee}:
        \vspace{-6mm}
        \begin{itemize}
            \item Word n-grams plus features such as typed-dependency relations and scores based on sentiment lexicons.
            \vspace{-2.5mm}
            \item $55.9\%$ best F$_1$ on the \textsc{data-bully} dataset.
        \end{itemize}
        \vspace{-2.5mm}
        Salminen et al. \shortcite{salminen2018anatomy}:
        \vspace{-2.5mm}
        \begin{itemize}
            \item \textsc{tf-idf} weighted n-grams constituted the best features.
            \vspace{-6mm}
            \item $79\%$ macro-F$_1$ using an \textsc{svm} on \textsc{data-ytube-fb}.
        \end{itemize}
        \vspace{-5mm}
        }\\ \hline
         
        \multicolumn{1}{p{7.5cm}|}{Advantages noted by Wiegand et al. \shortcite{wiegand} are that hand-crafted rules and lexicons generalize well across data from different domains, and that they can capture targeted aspects which are not easily learnable just from labeled data.}
        & 
        \multicolumn{1}{|p{7.5cm}}{Advantages include faster modeling since lesser human inputs are required in comparison. Moreover, features like character n-grams are robust to spelling, punctuation and grammatical variations \cite{nobata}.}\\ \hline
        
        \multicolumn{1}{p{7.5cm}|}{Disadvantages noted by Spertus \shortcite{smokey} and Wiegand et al. \shortcite{wiegand} included the inability of rules and lexicons to deal with implicit abuse and sarcasm, and their vulnerability to errors in spelling, punctuation and grammar.}
        & 
        \multicolumn{1}{|p{7.5cm}}{Disadvantages include features being relatively difficult to interpret directly. Additionally, features generated by this approach may only capture surface-level patterns but not deeper semantic properties \cite{wiegand}.}\\
        \bottomrule
    \end{tabular}
    \caption{The two approaches to textual feature engineering. First row discusses some of the works across the two, and second and third rows highlight some advantages and disadvantages of the two respectively.}
    \label{tab:text_feature_engg}
\end{table}

\noindent
\textbf{Social feature engineering.} Several researchers have directly incorporated features and identity traits of users in order to model the likeliness of abusive behavior from users with certain traits, a process known as \textit{user profiling}. Dadvar et al. \shortcite{davdar} included the age of users alongside other traditional lexicon-based features to detect cyber-bullying, while Gal{\'a}n-Garc{\'\i}a et al. \shortcite{galan2016supervised} utilized the time of publication, geo-position and language in the profile of Twitter users. Waseem and Hovy \shortcite{waseem_hovy} exploited gender of Twitter users alongside character n-gram counts to improve detection of sexism and racism in tweets from \textsc{data-twitter-wh} (F$_1$ increased from $73.89\%$ to $73.93\%$). Using the same setup, Unsv\aa g and Gamb\"ack \shortcite{unsvaag2018effects} showed that the inclusion of social network-based (i.e., number of followers and friends) and activity-based (i.e., number of status updates and favorites) information of users alongside their gender further enhances performance ($3$ points gain in F$_1$).


\section{Neural networks based abuse detection}
Advancements in computational capabilities have led researchers to explore methods for abuse detection that rely on neural architectures. Such methods can be broadly divided into three categories: 1) those that simply consume distributed representations generated by neural networks, 2) those that perform deep learning on texts, and 3) those that use neural networks for modeling social aspects.

\noindent
\textbf{Distributed representations.} Djuric et al. \shortcite{djuric} were the first to adopt neural networks for abuse detection. They utilized \textit{paragraph2vec} \cite{paragraph2vec} to obtain low-dimensional representations for comments in \textsc{data-yahoo-fin-dj} and trained a logistic regression (LR) classifier. Their model outperformed other classifiers trained on BOW-based representations (AUROC $80.07\%$ vs. $78.89\%$). The authors noted that words and phrases in hate speech tend to be obfuscated, leading to high dimensionality and sparsity of BOW-based representations, which in turn causes classifiers to over-fit in training.

Building on the work of Djuric et al., Nobata et al. \shortcite{nobata} examined the performance of a variety of features on the \textit{Yahoo!} datasets (\textsc{data-yahoo-*}) using a regression model: 1) word and character n-grams, 2) linguistic features like number of polite/hate words and punctuation count, 3) syntactic features like parent and grandparent of node in a dependency tree, and 4) distributional-semantic features like \textit{paragraph2vec} comment representations. Although the best results were achieved with all the features combined ($79.5\%$ F$_1$ on \textsc{data-yahoo-fin-a}, $81.7\%$ F$_1$ on \textsc{data-yahoo-news-a}), character n-grams on their own contributed significantly more than the other features due to their robustness to noise such as obfuscations, misspellings, unseen words. The  \textit{paragraph2vec} representations, in comparison, performed at par with character n-grams only on \textsc{data-yahoo-news-a}, which was noted to be less noisy than \textsc{data-yahoo-fin-a}. Working with the \textsc{data-yahoo-fin-dj} dataset, Mehdad and Tetreault \shortcite{mehdad} investigated whether character-level features are more indicative of abuse than word-level ones. Their experiments demonstrated the superiority of character-level features, revealing that SVM classifiers trained on Bayesian log-ratio vectors of average counts of character n-grams improve not only upon the more intricate approach of Nobata et al. (AUROC from $91\%$ to $92\%$), but also upon other recurrent neural networks (RNN) based character and word-level models.

Samghabadi et al. \shortcite{W17-3010} started with a similar set of features as Nobata et al. \shortcite{nobata} and augmented it with hand-engineered ones such as polarity scores derived from \href{http://sentiwordnet.isti.cnr.it}{\textit{SentiWordNet}}, scores based on the \href{http://liwc.wpengine.com}{\textit{LIWC}} program, and features based on emoticons. They applied their method to three different datasets: \textsc{data-wiki-att}, a \textit{Kaggle} dataset annotated for insult, and a dataset of questions and answers (each labeled as \textit{invective} or \textit{neutral}) that they created by crawling \textit{ask.fm}. Distributional-semantic features combined with the aforementioned features constituted an effective feature space for the task ($65\%$, $68\%$, $56\%$ F$_1$ on \textsc{data-wiki-att}, \textit{Kaggle}, \textit{ask.fm} respectively). In line with the results of Nobata et al. and Mehdad and Tetreault, the authors found character n-grams to be performing well on these datasets too. 

\vspace{1mm}
\noindent
\textbf{Deep learning on texts.} With the advent of deep learning, many researchers have explored its efficacy in abuse detection. Badjatiya et al. \shortcite{badjatiya} evaluated several neural architectures on the \textsc{data-twitter-wh} dataset. Their best setup involved a two-step approach wherein they used a 
word-level long-short term memory (LSTM) model, 
to tune GLoVe or randomly-initialized word embeddings, and then trained a gradient-boosted decision tree (GBDT) classifier on the average of the tuned embeddings in each tweet. They achieved the best results using randomly-initialized embeddings (weighted F$_1$ of $93\%$). However, working with a similar setup, Mishra et al. \shortcite{mishra} reported that GLoVe initialization provided superior performance; a mismatch was attributed to the fact that Badjatiya et al. tuned the embeddings on the entire dataset, including the test set, to which the randomly-initialized ones overfit better.

Park and Fung \shortcite{park_fung} utilized character and word-level CNNs 
to classify comments in the dataset that they formed by combining \textsc{data-twitter-w} and \textsc{data-twitter-wh}. Their experiments demonstrated that combining the two levels of granularity using two input channels achieves the best results, improving upon a character n-grams based LR baseline (weighted F$_1$ from $81.4\%$ to $82.7\%$). 
Several other works have also demonstrated the efficacy of CNNs in detecting abusive social media posts \cite{W18-5106}. Some researchers \cite{W18-5111,ziqizhang} have shown that sequentially combining CNNs with gated recurrent unit (GRU) RNNs can enhance performance by taking advantage of properties of both architectures, e.g., 1-2\% increase in F$_1$ compared to only using CNNs.

Pavlopoulos et al. \shortcite{pavlopoulos,pavlopoulos-emnlp} applied deep learning to the \textsc{data-wiki-att}, \textsc{data-wiki-tox}, and \textsc{data-gazzetta} datasets. Their most effective setups were: 1) a word-level GRU followed by an LR layer,  
and 2) setup 1 extended with an attention mechanism on words. 
Both setups outperformed a simple word-list baseline as well as the character n-grams based LR classifier (\textsc{detox}) from Wulczyn et al. \shortcite{wulczyn}. Setup 1 achieved the best performance on \textsc{data-wiki-att} (AUROC $97.71\%$) and \textsc{data-wiki-tox} (AUROC $98.42\%$), while setup 2 performed the best on \textsc{data-gazzetta} (AUROC $84.69\%$). Additionally, the attention mechanism was shown to be able to highlight abusive words and phrases within the comments, exhibiting a high level of agreement with annotators on the task. 
Lee et al. \shortcite{W18-5113} worked with a subset of the \textsc{data-twitter-f} dataset and 
showed that a word-level bi-GRU model along with \textit{latent topic clustering}, where topic information is extracted from the hidden states of the GRU \cite{N18-1142}, yielded the best weighted F$_1$ of $80.5\%$.

The \textit{GermEval 2018} shared task on identification of offensive language in German tweets \cite{wiegand2018overview} saw submission of both deep learning and feature engineering based methods. The winning system \cite{tuwien}, with macro F$_1$ of $76.77\%$, employed multiple character and token n-gram classifiers alongside distributional semantic features obtained by averaging word embeddings. The second best approach \cite{spmmmp}, with macro F$_1$ $75.52\%$, on the other hand, employed an ensemble of CNNs whose outputs were fed to a meta classifier for final prediction. 
Most of the remaining submissions \cite{hpitm,udsw} used deep learning with CNNs and RNNs alongside techniques such as transfer learning (e.g., via machine translation or joint representation learning for words across languages) from abuse-annotated datasets in other languages (mainly English). Wiegand et al. \shortcite{wiegand2018overview} noted that simple deep learning approaches themselves were quite effective, and the addition of other techniques did not necessarily provide substantial gains. 

Kumar et al. \shortcite{kumar2018benchmarking} noted similar trends in the shared task for aggression identification on the \textsc{data-facebook} dataset. The top approach on the task's English dataset \cite{aroyehun2018aggression}, with macro F$_1$ of $64.25\%$, comprised RNNs and CNNs along with transfer learning via machine translation. The top approach for Hindi \cite{samghabadi2018ritual}, with F$_1$ of $62.92\%$, utilized lexical features based on word and character n-grams. In order to further understand the pros and cons of both, Aken et al. \shortcite{van2018challenges} performed a systematic comparison of neural and non-neural approaches to toxic comment classification, concluding that ensembles of the two were most effective. 

The \textit{GermEval 2019} shared task on identification of offensive language in German tweets \cite{germeval2019} consisted of three sub-tasks: 1) course-grained classification of samples as \textit{offense} or other, 2) fine-grained classification of samples as \textit{offense}, \textit{profanity}, \textit{insult} or \textit{other}, and 3) classification of offensive samples as \textit{implicit} or \textit{explicit}. All three sub-tasks saw submission of a range of methods, including those based on deep contextualized language models like BERT \cite{bert}. The organizers noted that in fact the winning submissions across all three sub-tasks utilized some form of BERT. Paraschiv and Cercel \shortcite{upb-germeval19} made the winning submission on sub-tasks 1 (76.95\% F$_1$) and 2 (53.59\% average F$_1$) that comprised a BERT model pre-trained on German Wikipedia and German Twitter corpora prior to being fine-tuned on the sub-task datasets. Risch et al. \shortcite{dedis-germeval19} had the winning submission on sub-task 3 (53.93\% F$_1$) that again comprised a BERT model pre-trained on German texts. While the organizers noted that methods based on traditional CNN and RNN models didn't feature in the top 3 on any sub-task, they found that ensemble models trained on character and token n-grams \cite{tuwienkbs-germeval19} and lexicon-based features \cite{fosil-germeval19} fared well.

Researchers have recently started exploring multi-task learning with neural networks for the purpose of abuse detection. Rajamanickam et al. \shortcite{santosh} demonstrated that jointly learning over emotion classification and abuse detection tasks leads to better performance on the latter. Detecting the affective nature of comments (e.g., disgust, anger, joy, fear, optimism) helps to detect abuse more accurately on Twitter posts, achieving an F$_1$ of 79.55 and 76.03 on \textsc{data-twitter-wh} and \textit{OffensEval} respectively. Samghabadi et al. \shortcite{samghabadi2019attending} utilize an emotion-aware attention mechanism, achieving a macro-F$_1$ of 83.56 on \textit{Kaggle} and 88.27 on \textsc{data-wiki-att}.  

\vspace{1mm}
\noindent
\textbf{Modeling social aspects with neural networks.} More recently, researchers have employed neural networks to extract representations or \textit{profiles} for users instead of manually leveraging traits like gender, location, etc. as discussed before. Working with the \textsc{data-gazzetta} dataset, Pavlopoulos et al. \shortcite{W17-4209} incorporated user embeddings into Pavlopoulos' setup 1 \shortcite{pavlopoulos,pavlopoulos-emnlp} described above. They divided all the users whose comments are included in \textsc{data-gazzetta} into 4 \textit{types} based on proportion of abusive comments (e.g., \textit{red} users if $>10$ comments and $\geq 66\%$ abusive comments), \textit{yellow} (users with $>10$ comments and $33\%-66\%$ abusive comments), \textit{green} (users with $>10$ comments and $\leq 33\%$ abusive comments), and \textit{unknown} (users with $\leq 10$ comments).  They then assigned unique randomly-initialized embeddings to users and added them as additional input to the LR layer alongside representations of comments obtained from the GRU. This increased the AUROC from $79.24\%$ to $80.71\%$. 
Qian et al. \shortcite{N18-2019} used LSTMs for modeling inter and intra-user relationships on \textsc{data-twitter-wh}, with sexist and racist tweets combined into one category. 
The authors applied a bi-LSTM to users' recent tweets in order to generate intra-user representations that capture their historic behavior. To improve robustness against noise present in tweets, they also used locality sensitive hashing to form sets of semantically similar user tweets. The authors then trained a policy network to select tweets from sets that a bi-LSTM could use to generate inter-user representations. When these inter and intra-user representations were utilized alongside representations of tweets from an LSTM baseline, F$_1$ increased from $70.3\%$ to $77.4\%$.

Mishra et al. \shortcite{mishra} constructed a community graph of all the users whose tweets are in the \textsc{data-twitter-wh} dataset. Nodes were the users and edges denoted the follower-following relationship among them on \textit{Twitter}. The authors applied \textit{node2vec} \cite{node2vec-kdd2016} to this graph to generate user embeddings, i.e., profiles. Inclusion of these embeddings into the character n-gram based baselines yielded significant gains on \textsc{data-twitter-wh} whereby F$_1$ scores on the racism and sexism classes increased from $72.28\%$ and $72.09\%$ to $75.09\%$ and $82.75\%$ respectively. The gains were attributed to the fact that user embeddings captured not only information about online communities, but also elements of the wider conversation amongst connected users. Ribeiro et al. \shortcite{ribeiro} and Mishra et al. \shortcite{mishragcn} applied graph neural networks \cite{gcn,graphsage} to community graphs to generate embeddings for users that capture not only their surrounding community but also their linguistic behavior. Mishra et al. \shortcite{mishragcn} recorded $79.49\%$ F$_1$ on the racism and $84.44\%$ F$_1$ on the sexism classes of \textsc{data-twitter-wh}.


\section{Discussion}
\subsection{Current trends and outstanding challenges in modeling abuse}
English has been the dominant language so far in terms of focus, followed by German, Hindi and Dutch. However, recent efforts have focused on compilation of datasets in other languages such as Slovene and Croatian \cite{slovene}, Chinese \cite{su2017}, Arabic \cite{mubarak-et-al-2017}, and even some unconventional ones such as \textit{Hinglish} \cite{hinglish}. Most of the research to date has been on racism, sexism, personal attacks, toxicity, and harassment. Other types of abuse such as obscenity, threats, insults, and grooming remain relatively unexplored. That said, we note that the majority of methods investigated to date and described herein are (in principle) applicable to a range of abuse types.

The recent approaches that rely on word-level CNNs and RNNs remain vulnerable to obfuscation of words \cite{mishra2}. On the other hand, the use of sub-word units, both in feature engineering (e.g. character n-grams) and as tokenized inputs to models like BERT, remains one of the most effective techniques for addressing obfuscation since sub-word units are robust to spelling variations. Many researchers to date have exclusively relied on text based features for abuse detection. But recent works have shown that personal and community-based profiling features of users significantly enhance the state of the art. Since posts on social media often includes data of multiple modalities (e.g., a combination of images and text), abuse detection systems would also need to incorporate a multi-modal component. Facebook recently released a dataset consisting of multi-modal hateful memes \cite{hate-memes} under the \textit{Hateful Memes Challenge} to foster research in this area.

Despite the fast-paced progress in this field, an important challenge that remains mostly unsolved is that of recognizing implicit abuse \cite{van2018challenges}. Implicit abuse comes in the form of figurative language, such as sarcasm, irony or metaphor, rhetorical questions, analogies and comparisons. Metaphor and sarcasm are particularly common, and tend to express stronger emotions and sentiments than the literally-used words and phrases \cite{Mohammad2016}. Nobata et al. \shortcite{nobata} (among others) noted that sarcastic comments are hard for abuse detection methods to deal with since surface features are not sufficient; typically the knowledge of the context or background of the user is also required. 
Mishra \shortcite{mishra_thesis} found that metaphors are more frequent in abusive samples as opposed to non-abusive ones. 
However, to fully understand the impact of figurative devices on abuse detection, datasets with more pronounced presence of these are required.
 
The key to modeling implicit abuse, and detecting abuse more accurately in general, might lie in shifting focus from modeling individual comments to modeling online conversations and how they evolve and escalate towards abuse. Abuse is inherently contextual; it can only be interpreted as part of a wider conversation between users on the Internet. This means that, in practice, individual comments can be difficult to classify without modeling their respective contexts. 
 Mishra et al. \shortcite{mishra} have pointed out that some tweets in \textsc{data-twitter-wh} do not contain sufficient lexical or semantic information to detect abuse even in principle, e.g., \textit{@user: Logic in the world of Islam http://t.co/xxxxxxx}, and techniques for modeling discourse and elements of pragmatics are needed. 
To address this issue, Gao and Huang \shortcite{gao2017detecting}, working with \textsc{data-fox-news}, incorporate features from two sources of context: the title of the news article for which the comment was posted, and the screen name of the user who posted it. Yet this is only a first step towards modeling the wider context in abuse detection; more sophisticated techniques are needed to capture the history of the conversation and the behavior of the users as it develops over time. NLP techniques for modeling discourse and dialogue can be a good starting point in this line of research. 


Another challenge in modeling abuse is presented by its ever-changing nature, as societies and technologies evolve. New abusive words and phrases continue to enter the language \cite{wiegand}. 
 Working with the \textsc{data-yahoo-*-b} datasets, Nobata et al. \shortcite{nobata} found that a classifier trained on more recent data outperforms one trained on older data. They noted that a prominent factor in this is the continuous evolution of the Internet jargon. We would like to add that, given the \textit{situational} and topical nature of abuse \cite{chandrasekharan2018internet}, contextual features learned by detection methods may become irrelevant over time. A similar trend also holds for abuse detection across domains. Wiegand et al. \shortcite{wiegand} showed that the performance of the (then) state of the art classifiers \cite{nobata,pavlopoulos-emnlp} decreases substantially when tested on data drawn from domains different to that of the training set. They attributed this trend to lack of domain-specific learning. 
Chandrasekharan et al. \shortcite{chandrasekharan2017bag} propose an approach that utilizes similarity scores between posts to improve in-domain performance based on out-of-domain data. 
Possible solutions for improving cross-domain abuse detection can be found in the literature of (adversarial) multi-task learning and domain adaptation \cite{daume2009frustratingly,ganin2016domain,wu2015collaborative}, and also in works such as that of 
Sharifirad et al. \shortcite{jafarpour2018boosting} who utilize knowledge graphs to augment the training of a sexist tweet classifier. Recently, Waseem et al. \shortcite{waseem2018bridging} and Karan and {\v{S}}najder \shortcite{karan2018cross} exploited multi-task learning frameworks to train models that are robust across data from different distributions or data annotated under different guidelines.  

\subsection{Ethical questions around automated abuse detection}
 Identifying experiences as abusive provides validation to victims of abuse and enables observers to grasp the scope of the problem. It also creates new descriptive norms, suggesting what types of behavior constitute abuse, and outlines existing expectations around appropriate behavior. On the other hand, automated systems can invalidate abusive experiences, particularly for victims whose experiences may not lie in the realm of \textit{typical} ones \cite{blackwell2017classification}. This points to a critical issue: automated systems embody the morals and values of their creators and annotators \cite{bowker2000sorting,blackwell2017classification}. It is therefore imperative that we design systems that are robust to such issues, e.g., some recent works have investigated ways to mitigate gender bias in models \cite{binns2017like,park2018reducing}.

That said, unfortunately, whilst the research community has started incorporating signals from user profiling, there has not yet been a discussion of ethical guidelines for doing so. To encourage such a discussion, we lay out four ethical considerations in the design of such approaches:
\begin{itemize}
    \item The profiling approach should not compromise the \textit{privacy} of the user. So a researcher might ask themselves such questions as: is the profiling based on identity traits of users (e.g., gender, race etc.) or solely on their online behavior? And is an appropriate generalization from (identifiable) user traits to population-level behavioral trends performed?
    \item One needs to reflect on the possible \textit{bias} in the training procedure; is the approach likely to induce a bias against users with certain traits?
    \item The \textit{visibility} aspect needs to be accounted for; is the profiling visible to the users, i.e., can users directly or indirectly observe how they (or others) have been profiled?
    \item One needs to carefully consider the \textit{purpose} of such profiling; is it intended to take actions against users, or is it more benign (e.g. to better understand the content produced by them and make task-specific generalizations)?
\end{itemize}

\noindent
While we do not intend to provide answers to these questions within this paper, we hope that the above considerations can help to start a debate on these important issues.

\subsection{Explainable abuse detection}
\textit{Explainability} has become an important aspect within NLP, and within AI generally. Yet there has been no discussion of this issue in the context of abuse detection systems. 
We hereby propose three properties that an \textit{explainable} abuse detection system should aim to exhibit.
\begin{itemize}
    \item It needs to \textit{establish intent} of abuse (or the lack of it) and provide evidence for it, hence convincingly segregating abuse from other phenomena such as sarcasm and humor.
    \item It needs to \textit{capture abusive language}, i.e., highlight instances of abuse if present, be they explicit (i.e., use of expletives) or implicit (e.g., dehumanizing comparisons).
    \item It needs to  \textit{identify the target(s)} of abuse (or the absence thereof), be it an individual or a group.
\end{itemize}

\noindent
These properties align well with the categorizations of abuse we discussed in the introduction. They also aptly motivate the advances needed in the field: (1) developments in areas such as sarcasm detection and user profiling for precise segregation of abusive intent from humor, satire, etc.; (2) better identification of implicit abuse, which requires improvements in modeling of figurative language; (3) effective detection of generalized abuse and inference of target(s), which require advances in areas such as domain adaptation and conversation modeling.

\section{Conclusions}
Online abuse stands as a significant challenge before society. Its nature and characteristics constantly evolve, making it a complex phenomenon to study and model. Methods for automated abuse detection have seen a lot of development in recent years: from simple rule-based ones aimed at identifying directed and explicit abuse to sophisticated ones that can capture rich semantic information and even aspects of user behavior. By providing a comprehensive review of the field to date, our paper aims to lay a platform for future research, facilitating progress in this important effort. While we see an array of challenges that lie ahead, e.g., modeling extra-propositional aspects of language, user behavior and wider conversation, we believe that recent progress in the areas of semantics, dialogue modeling and social media analysis put the research community in a strong position to address them. The notion of abuse has been rather hard to define due to differing opinions on sarcasm, self-deprecating humor, and terms seen as offensive, to name a few issues. In fact, attempts to impose a definition may also curb the diversity that exists across various abuse datasets since what constitutes abuse varies from culture to culture. But we do believe that the NLP community can and should work towards standardizing the understanding of different characteristics of abuse, examples of which are presented in the paper: directed, generalized, implicit and explicit. This will allow for more comparable and systematic modeling of different types of abuse (including those that might emerge in the future) and also facilitate transfer learning across them.





\bibliographystyle{coling}
\bibliography{coling2020}

\clearpage
\begin{appendices}
\section{Summaries of public datasets}
In table \ref{dataset_urls}, we summarize the datasets described in this paper that are publicly available and provide links to them.

\section{A discussion of metrics}
The performance results we have reported highlight that, throughout work on abuse detection, different researchers have utilized different evaluation metrics for their experiments -- from area under the receiver operating characteristic curve (AUROC) \cite{wulczyn,djuric} to micro and macro F$_1$ \cite{mishra2} -- regardless of the properties of their datasets. This makes the presented techniques more difficult to compare. In addition, as abuse is a relatively infrequent phenomenon, the datasets are typically skewed towards non-abusive samples \cite{waseem}. Metrics such as AUROC may, therefore, be unsuitable since they may mask poor performance on the abusive samples as a side-effect of the large number of non-abusive samples \cite{metrics}. \textit{Macro-averaged} precision, recall, and F$_1$, as well as precision, recall, and F$_1$ on specifically the abusive classes, may provide a more informative evaluation strategy; the primary advantage being that macro-averaged metrics provide a sense of effectiveness on the minority classes \cite{van2013macro}. Additionally, area under the precision-recall curve (AUPRC) might be a better alternative to AUROC in imbalanced scenarios \cite{prc}.

\begin{table}[ht]
\centering
\begin{adjustbox}{width=23cm, angle=270}
\begin{tabular}{| c | c | c | c | c | c |}
\hline
\textbf{Dataset link} & \textbf{Associated paper} & \textbf{Language} & \textbf{Size} & \textbf{Source} & \textbf{Composition}\\ \hline
\href{http://github.com/zeerakw/hatespeech}{\textsc{data-twitter-wh}} & \cite{waseem_hovy} & English & $17k$ & \textit{Twitter} & \textit{Racism}, \textit{Sexism}, \textit{Neither}\\ \hline
\href{http://github.com/zeerakw/hatespeech}{\textsc{data-twitter-w}} & \cite{waseem} & English & $7k$ & \textit{Twitter} & \textit{Racism}, \textit{Sexism}, \textit{Both}, \textit{Neither}\\ \hline
\href{https://github.com/t-davidson/hate-speech-and-offensive-language}{\textsc{data-twitter-david}} & \cite{davidson} & English & $25k$ & \textit{Twitter} & \textit{Racist}, \textit{Offensive}, \textit{Clean}\\ \hline
\href{https://github.com/ENCASEH2020/hatespeech-twitter}{\textsc{data-twitter-f}} & \cite{founta} & English & $80k$ & \textit{Twitter} & \textit{Abusive}, \textit{Spam}, \textit{Hateful}, \textit{Normal}\\ \hline
\href{https://meta.wikimedia.org/wiki/Research:Detox/Data_Release}{\textsc{data-wiki-att}} & \cite{Wulczyn2017} & English & $116k$ & \textit{Wikipedia talk page} & \textit{Personal attack}, \textit{Clean}\\ \hline
\href{https://meta.wikimedia.org/wiki/Research:Detox/Data_Release}{\textsc{data-wiki-agg}} & \cite{Wulczyn2017} & English & $116k$ & \textit{Wikipedia talk page} & \textit{Aggressive}, \textit{Clean}\\ \hline
\href{https://meta.wikimedia.org/wiki/Research:Detox/Data_Release}{\textsc{data-wiki-tox}} & \cite{Wulczyn2017} & English & $160k$ & \textit{Wikipedia talk page} & \textit{Toxic}, \textit{Clean}\\ \hline
\href{https://github.com/sjtuprog/fox-news-comments}{\textsc{data-fox-news}} & \cite{gao2017detecting} & English & $1.5k$ & \textit{Fox news} & \textit{Hateful}, \textit{Non-hateful}\\ \hline
\href{http://nlp.cs.aueb.gr/software.html}{\textsc{data-gazzetta}} & \cite{pavlopoulos-emnlp} & Greek & $1.6M$ & \textit{Gazzetta} & \textit{Accept}, \textit{Reject}\\ \hline
\href{http://trac1-dataset.kmiagra.org/}{\textsc{data-facebook}} & \cite{kumar2018benchmarking} & Hindi \& English & $15k$ & \textit{Facebook} & \textit{\{Covertly, Overtly, Non\}-aggressive}\\ \hline
\href{https://www.kaggle.com/c/detecting-insults-in-social-commentary/data}{\textit{Arabic News}} & \cite{mubarak-et-al-2017} & Arabic & $32k$ & \textit{Aljazeera News} & \textit{Obscene}, \textit{Offensive}, \textit{Clean}\\ \hline
\href{https://github.com/uds-lsv/GermEval-2018-Data}{\textit{GermEval 2018}} & \cite{wiegand2018overview} & German & $8.5k$ & \textit{Twitter} & \textit{Abuse}, \textit{Insult}, \textit{Profanity}, \textit{Other}\\ \hline
\href{http://ritual.uh.edu/resources/}{\textit{Ask.fm}} & \cite{W17-3010} & English & $5.6k$ & \textit{Ask.fm} & \textit{Invective}, \textit{Neutral}\\ \hline
\end{tabular}
\end{adjustbox}
\caption{Links and summaries of datasets mentioned in the paper that are publicly available.}
\label{dataset_urls}
\end{table}

\section{Embeddings and OOV words}
Djuric et al. \shortcite{djuric} and Nobata et al. \shortcite{nobata} observed that abusive language tends to contain obfuscations (e.g., \textit{w0m3n}). This poses a particular problem for abuse detection methods \cite{blackwell2017classification} and specifically for those that rely on word-level neural networks as they operate with a finite vocabulary of words and map all unknown words in the test set to a single out-of-vocabulary (OOV) embedding. This has the undesired effect that deliberately obfuscated words and benign misspellings get conflated, leading to loss in performance \cite{mishra,N18-2019}. Spelling correction and edit-distance techniques for resolving obfuscations can provide some level of mitigation; however, they do not help in cases where obfuscation is severe, e.g., \textit{a55h0le}, or is by concatenation, e.g., \textit{idiotb*tch}. While one way around the problem is to have character-level models, Mishra et al. \shortcite{mishra2} showed that such models perform worse than word-level ones with pre-trained embeddings. Hence, techniques to generate embeddings for OOV words on the fly \cite{bojanowski,mishra2} have been exploited.
\end{appendices}

\end{document}